\documentclass[runningheads]{llncs}
\usepackage{graphicx}
\usepackage{comment}
\usepackage{amsmath,amssymb} 
\usepackage{color, colortbl}

\usepackage{epsfig}
\usepackage{graphicx}
\usepackage{amssymb}
\usepackage{xcolor}
\usepackage{nicefrac}
\usepackage{booktabs}
\usepackage{placeins}
\usepackage{pifont}
\usepackage{subcaption}
\usepackage{xspace}
\usepackage{arydshln}
\usepackage[pagebackref=true,breaklinks=true,letterpaper=true,colorlinks,bookmarks=false]{hyperref}

\newcommand{\mypm}{\,$\pm$\,}

\definecolor{BrickRed}{HTML}{B6321C}
\definecolor{RoyalBlue}{HTML}{0071BC}
\definecolor{PineGreen}{HTML}{008B72}
\definecolor{bluefig}{HTML}{5B9BD5}
\definecolor{Gray}{gray}{0.9}

\newcommand*\OK{\ding{51}}

\let\originalparagraph\paragraph
\renewcommand{\paragraph}[2][.]{\originalparagraph{#2#1}}

\makeatletter
    \newcommand{\fs}{\@ifnextchar.{}{.}}
\makeatother

\let\thetaold\theta
\renewcommand{\theta}{\boldsymbol{\thetaold}}
\newcommand{\mcL}{\mathcal{L}}
\newcommand{\vx}{\mathbf{x}}
\newcommand{\vh}{\mathbf{h}}
\newcommand{\vy}{\mathbf{y}}
\newcommand{\vyh}{\hat\vy}
\newcommand{\parag}[1]{\vspace{0.2cm}\noindent\textbf{#1}.\ }
\newcommand{\pp}{\,\textit{p.p}\fs}

\begin{document}
\pagestyle{headings}
\mainmatter
\def\ECCVSubNumber{3384}  


\title{PODNet: Pooled Outputs Distillation for Small-Tasks Incremental Learning}
\titlerunning{PODNet: Pooled Outputs Distillation}
%
\author{Arthur Douillard\inst{1,2} \and
Matthieu Cord\inst{2,3} \and
Charles Ollion\inst{1} \and
Thomas Robert\inst{1} \and 
Eduardo Valle\inst{4}
}
%
%
\institute{Heuritech, Paris, France
\\\email{\{arthur.douillard,thomas.robert,charles.ollion\}@heuritech.com} \and
Sorbonne University, Paris, France \\\email{matthieu.cord@sorbonne-universite.fr}
\\
\and valeo.ai, Paris, France\\
 \and
 University  of Campinas, Campinas, Brazil\\
\email{dovalle@dca.fee.unicamp.br}}
\authorrunning{A. Douillard et al.}
\maketitle

\begin{abstract}
Lifelong learning has attracted much attention, but existing works still struggle to fight catastrophic forgetting and accumulate knowledge over long stretches of incremental learning. 
In this work, we propose PODNet, a model inspired by representation learning. By carefully balancing the compromise between remembering the old classes and learning new ones, PODNet fights catastrophic forgetting, even over very long runs of small incremental tasks --\,a setting so far unexplored by current works. PODNet innovates on existing art with an efficient spatial-based distillation-loss applied throughout the model and a representation comprising multiple proxy vectors for each class. We validate those innovations thoroughly, comparing PODNet with three state-of-the-art models on three datasets: CIFAR100, ImageNet100, and ImageNet1000. Our results showcase a significant advantage of PODNet over existing art, with accuracy gains of 12.10, 6.51, and 2.85 percentage points, respectively.\footnote{Code is available at: \href{https://github.com/arthurdouillard/incremental_learning.pytorch}{\texttt{github.com/arthurdouillard/incremental\_learning.pytorch}}}

\end{abstract}

\section{Introduction}

Lifelong machine learning~\cite{robins1995catastrophicforgetting,french1999catastrophicforgetting,thrun1998lifelonglearning} focuses on models that accumulate and refine knowledge over large timespans. Incremental learning --\,the ability to aggregate different learning objectives seen over time into a coherent whole\,-- is paramount to those models. To achieve incremental learning, models must fight \textit{catastrophic forgetting}~\cite{robins1995catastrophicforgetting,french1999catastrophicforgetting} of previous knowledge. Lifelong and incremental learning have attracted much attention in the past few years, but existing works still struggle to preserve acquired knowledge over many cycles of short incremental learning steps. 

We will focus on image classifiers, which are ordinarily trained once on a fixed set of classes. In \textit{incremental learning}, however, the classifier must learn the classes by steps, in training cycles called \textit{tasks}. At each task, we expose the classifier to a new set of classes. Incremental learning would reduce trivially to ordinary classification if we were allowed to store all training samples, but we are imposed a limited \textit{memory}: a maximum number of samples for previously learned classes. This limitation is motivated by practical applications, in which privacy issues, or storage and computing limitations prevent us from simply retraining the entire model for each new task~\cite{li2018lwf,lomonaco2017core50}. Furthermore, incremental learning is different from transfer learning in that we aim to have good performance in both old and new classes.

To overcome catastrophic forgetting, different approaches have been proposed: reusing a limited amount of previous training data~\cite{rebuffi2017icarl,castro2018end_to_end_inc_learn}; learning to generate the training data~\cite{kemker2018fearnet,shin2017deep_generative_replay}; extending the architecture for new phases of data~\cite{yoon2018dynamically_expandable_networks,li2019learning_to_grow}; using a sub-network for each phase~\cite{fernando2017path_net,golkar2019neural_pruning}; or constraining the model divergence as it evolves~\cite{kirkpatrick2017ewc,lopezpaz2017gem,aljundi2018MemoryAwareSynapses,li2018lwf,rebuffi2017icarl,castro2018end_to_end_inc_learn}. 

In this work, we propose PODNet, approaching incremental learning as representation learning, with a distillation loss that constrains the evolution of the representation. By carefully balancing the compromise between remembering the old classes and learning new ones, we learn a representation that fights catastrophic forgetting, remaining stable over long runs of small incremental tasks. Our model innovates on existing art with (1) an \textit{efficient spatial-based} distillation-loss applied \textit{throughout the model}; and (2) as a refinement, a representation comprising multiple proxy vectors for each class, resulting in a more flexible representation.

In this paper, we first present the existing state of the art (\autoref{sec:related_work}), which we close by detailing our contributions. We then describe our model (\autoref{sec:model}), and evaluate it in an extensive set of experiments (\autoref{sec:expes}) on CIFAR100, ImageNet100, and ImageNet1000, including ablation studies assessing each contribution, and extensive comparisons with existing methods. 

\section{Related Work}
\label{sec:related_work}

To approach the problem of incremental learning, consider a single incremental task: one has a classifier already trained over a set of old classes and must adapt it to learn a set of new classes.
To perform that single task, we will consider: (1) the data/class representation model; (2) the set of constraints to prevent catastrophic forgetting; (3) the experimental context (including the constraints over the memory for previous training data) for which to design the model.

\parag{Data/class representation model} Representation learning was already implicitly present in iCaRL~\cite{rebuffi2017icarl}: it introduced the Nearest Mean Exemplars (NME) strategy which averages the outputs of the deep convolutional network to create a single proxy feature vector per class that are then used by a nearest-neighbor classifier predict the final classes. Hou et al.~\cite{hou2019ucir} adopted this method and also introduced another, named CNN, which uses the output class probabilities to classify incoming samples, freezing (during training) the classifier weights associated with old classes, and then fine-tuning them on an under-sampled dataset.

Hou et al.~\cite{hou2019ucir}, in the method called here UCIR, made representation learning explicit, by noticing that the limited memory imposed a severe imbalance on the training samples available for the old and for the new classes. To overcome that difficulty, they designed a metric-learning model instead of a classification model. That strategy is often used in few-shot learning~\cite{gidaris2018fewshot_wo_forgetting} because of its robustness to few data. Because classical metric architectures require special training sampling (e.g., semi-hard sampling for triplets), Hou et al. chose instead to redesign the classifier's last layer of their model to use the cosine similarity~\cite{luo2018cosine_classifier}.

\parag{Model constraints to prevent catastrophic forgetting} Constraining the model's evolution to prevent forgetting is a fruitful idea proposed by several methods~\cite{kirkpatrick2017ewc,lopezpaz2017gem,aljundi2018MemoryAwareSynapses,li2018lwf,rebuffi2017icarl,castro2018end_to_end_inc_learn}. 
Preventing the model's parameters from diverging too much forces it to remember the old classes, but care must be taken to still allow it to learn the new ones. We call this balance the \textit{rigidity-plasticity trade-off}.

Existing art on knowledge distillation/compression~\cite{hinton2015knowledge_distillation} was an important source of inspiration for constraints on models. The goal is to distill a large trained model (called teacher) into a new smaller model (called student). The distillation loss forces the features of the student to approach those of its teacher. In our case, the student is the current model and the teacher---with same capacity-- is its version at the previous task. Zagoruyko and Komodakis~\cite{komodakis2017attention_residual_distillation} investigated attention-based distillation for image classifiers, by pooling the intermediate features of convolutional networks into attention maps, then used in their distillation losses.
Li and Hoiem~\cite{li2018lwf} —-\,and several authors after them~\cite{rebuffi2017icarl,castro2018end_to_end_inc_learn,wu2019bias_correction}\,—- used a binary cross-entropy between the output probabilities by the models.
Hou et al.~\cite{hou2019ucir}, used instead \textit{Less-Forget}, a cosine-similarity constraint on the flat feature embeddings after the global average pooling.
Dhar et al.~\cite{dhar2019learning_without_memorizing_gradcam} proposed to constrain the gradient-based attentions generated by GradCam~\cite{selvaraju2017gradcam}, a visualization method.
Wu et al.~\cite{wu2019bias_correction} proposed BiC, an algorithm oriented towards large-scale datasets, which employs a small linear model learned on validation data to recalibrate the output probabilities before applying a distillation loss.

\begin{figure}[tb]
\begin{center}
    \includegraphics[width=0.8\linewidth]{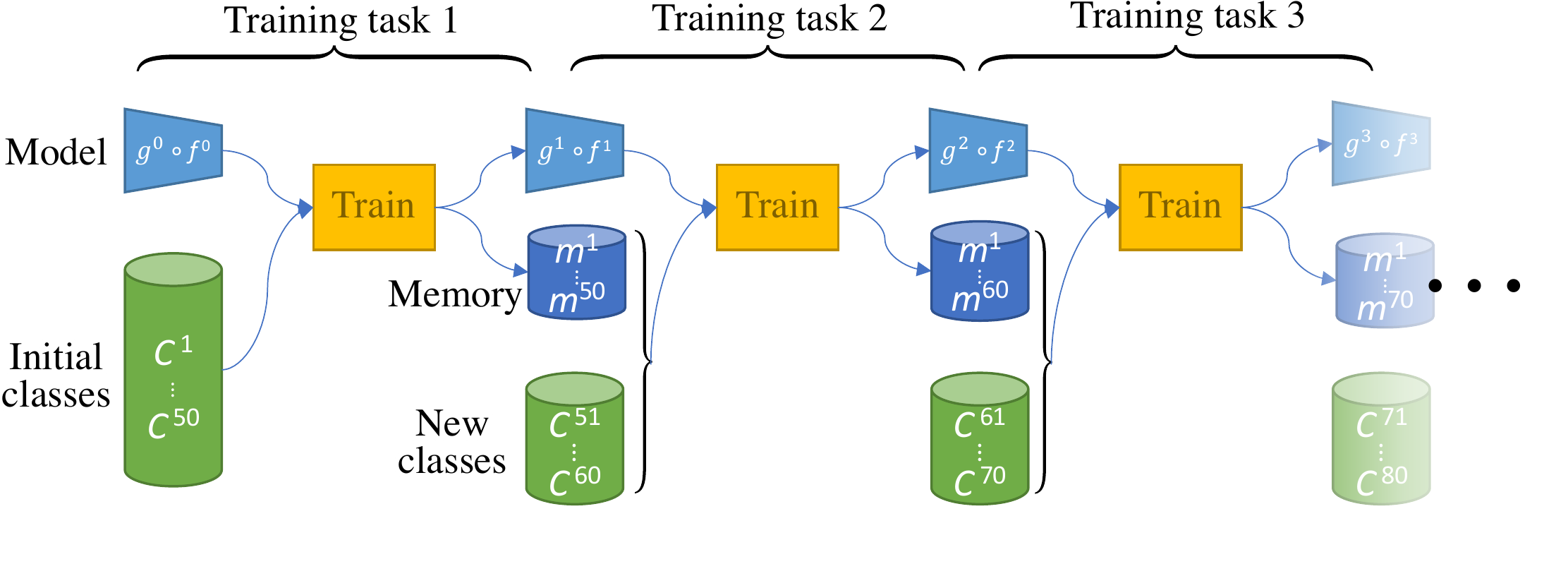}
\end{center}
   \caption{\textbf{Training protocol for incremental learning}. At each training task we learn a new set of classes, and the model must retain knowledge about \textit{all} classes. The model is allowed a \textit{limited} memory of samples of old classes.}
    \label{fig:protocol}
\end{figure}

\parag{Experimental context} A critical component of incremental learning is the convention used for the memory storing samples of previous data. An usual convention is to consider a fixed amount of samples allowed in that memory, as illustrated in \autoref{fig:protocol}.

Still, there are two experimental protocols for such fixed-sample convention: we may either use the memory budget at will ($M_\mathrm{total}$), or add a constraint on the number of samples per class for the old classes ($M_\mathrm{per}$). When $M_\mathrm{total}=M_\mathrm{per}\times$\textit{\# of classes}, both settings have equivalent \textit{final} memory size, but the latter, that we adopt, is much more challenging since early tasks cannot benefit from the full memory size.
\textit{The granularity of the increments} is another critical element: with a fixed number of classes, increasing the number of tasks decreases the number of classes per task. More tasks imply stronger forgetting of the earliest classes, and pushing that number creates a challenging protocol, so far unexplored by existing art. Hou et al. evaluate at most 10 tasks on CIFAR100, while we propose as much as 50 tasks.

Finally, to score the experiments, Rebuffi et al.~\cite{rebuffi2017icarl} proposed a global metric that they called \textbf{average incremental accuracy}, taking into account the entire history of the run, averaging the accuracy at the end of each task (including the first).

\parag{Contributions} As seen, associating representation learning to model constraints is a particularly fruitful idea for incremental learning, but requires carefully balancing the goals of rigidity (to avoid catastrophic forgetting) and plasticity (to learn new classes). 

Employing a distillation-based loss to constrain the evolution of the representation has also resulted in leading results~\cite{hou2019ucir,wu2019bias_correction,peng2019m2kd,dhar2019learning_without_memorizing_gradcam}. Our model improves existing art by employing a \textit{novel and efficient spatial-based} distillation loss, which we are able to apply \textit{throughout the model}.

Implicit or explicit proxy vectors representing each class inside the models have lead to state of the art results~\cite{rebuffi2017icarl,hou2019ucir}. Our model extends that idea allowing for \textit{multiple proxy vectors} per class, resulting in a more flexible representation.
\section{Model}
\label{sec:model}

Formally, we learn the model in $T$ \textit{tasks}, task $t$ comprising a set of new classes $C^t_N$, and a set of old classes $C^t_O$, and aiming at classifying all seen classes $C^t_O \cup C^t_N$. Between tasks, the new set $C^t_O$ will be set to $C^{t-1}_O \cup C^{t-1}_N$, but the amount of training samples from $C^t_O$ (called \textit{memory}) is constrained to exactly $M_\mathrm{per}$ samples per class, while all training samples in the dataset are allowed for the classes in $C^t_N$, as shown in \autoref{fig:protocol}. The resulting imbalance, if unmanaged, leads to \textit{catastrophic forgetting}~\cite{robins1995catastrophicforgetting,french1999catastrophicforgetting}, i.e., learning the new classes at the cost of forgetting the old ones.

\begin{figure}[t]
\begin{center}
    \includegraphics[width=0.8\linewidth]{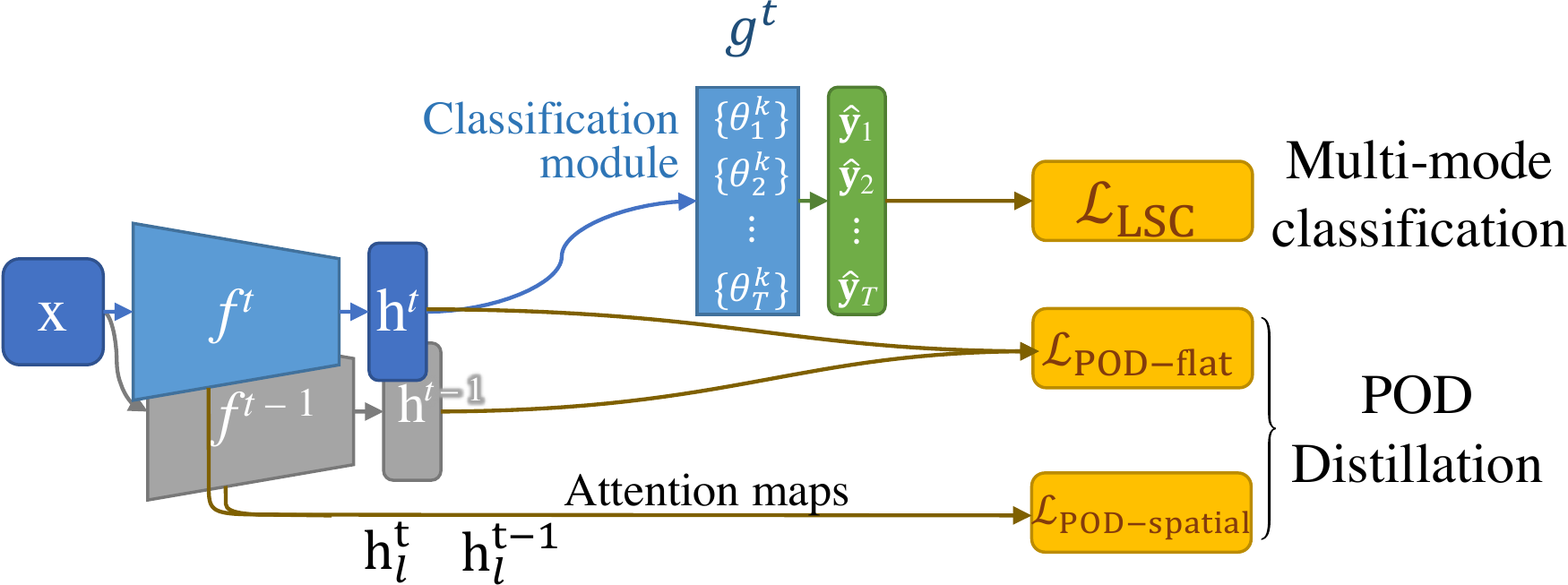}
\end{center}
   \caption{\textbf{Overview of PODNet}: the distillation loss POD prevent excessive model drift by constraining intermediate outputs of the ConvNet $f$ and the LSC classifier $g$ learns a more expressive multi-modal representation.}
    \label{fig:model}
\end{figure}

Our base model is a deep convolutional network $\vyh = g(f(\vx))$, where $\vx$ is the input image, $\mathbf{y}$ is the output vector of class probabilities, $\vh = f(\vx)$ is the ``feature extraction'' part of the network (all layers up to the next-to-last), $\vyh = g(\vh)$ is the final classification layer, and $\vh$ is the final embedding of the network before classification (\autoref{fig:model}). The superscript $t$ denotes the model learned at task $t$:$f^{t}$, $g^{t}$, $\vh^{t}$, etc.





\subsection{POD: Pooled Outputs Distillation loss}
\label{sec:distillation}

Constraining the evolution of the weights is crucial to reduce forgetting. Each new task $t$ learns a new (student) model, whose weights are not only initialized with those of the previous (teacher) model, but also constrained by a distillation loss. That loss must be carefully balanced to prevent forgetting (rigidity), while allowing the learning of new classes (plasticity).

To this goal, we propose a set of constraints we call \textbf{Pooled Outputs Distillation (POD)}, applied not only over the final embedding output by $\vh^{t}=f^{t}(\vx)$, but also over the output of its intermediate layers $\vh^{t}_\ell=f^{t}_\ell(\vx)$ (where by notation overloading
$f^{t}_\ell(\vx)\equiv f^{t}_\ell\circ\ldots\circ f^{t}_1(\vx)$, and thus
$f^{t}(\vx)\equiv f^{t}_L\ldots\circ f^{t}_\ell\circ\ldots f^{t}_1(\vx)$).

\begin{figure}[tb]
\begin{center}
    \includegraphics[width=0.90\linewidth]{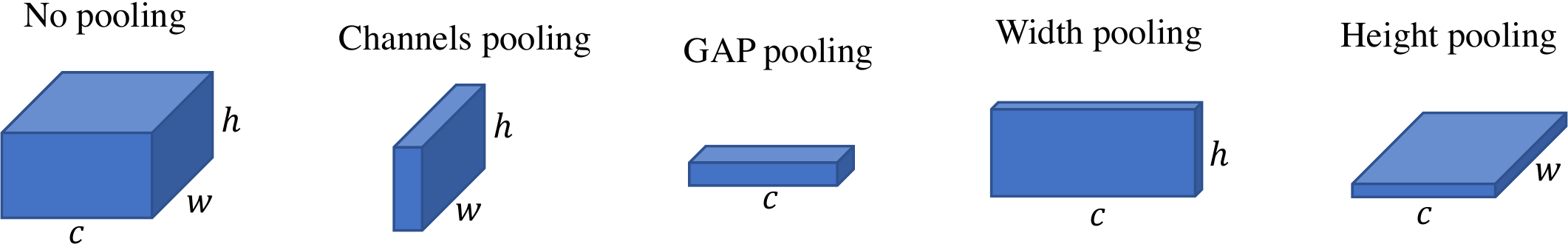}
\end{center}
   \caption{\textbf{Different possible poolings}. The output from a convolutional layer $\vh^{t}_{\ell,c,w,h}$ may be pooled (summed over) one or more axes. The resulting loss considers only the pooled activations instead of the individual components, allowing more plasticity across the pooled axes.}
    \label{fig:pooling}
\end{figure}

The convolutional layers of the network output tensors $\vh^{t}_{\ell}$ with components $\vh^{t}_{\ell,c,w,h}$, where $c$ stands for channel (filter), and $w\times h$ for column and row of the spatial coordinates. The loss used by POD may pool (sum over) one or several of those indexes, more aggressive poolings (\autoref{fig:pooling}) providing more freedom, and thus, plasticity: the lowest possible plasticity imposes an exact similarity between the previous and current model while higher plasticity relaxes the similarity definition.

Pooling is an important operation in Computer Vision, with a strong theoretical motivation. In the past, pooling has been introduced to obtain invariant representations~\cite{lowe1999sift,lazbnik2006spatial_pyramid_matching}. Here, the justification is similar, but the goal is different: as we will see, the pooled indexes are aggregated in the proposed loss, allowing \textit{plasticity}. Instead of the model acquiring invariance to the input image, the desired loss acquires invariance to model evolution, and thus, representation.
The proposed pooling-based formalism has two advantages: first, it organizes disparately proposed distillation losses into a neat, general formalism. Second, as we will see, it allowed us to propose novel distillation losses, with better plasticity-rigidity compromises. Those topics are explored next.

\parag{Pooling of convolutional outputs} As explained before, POD constrains the output of each intermediate convolutional layer $\vh^{t}_{\ell,c,w,h} = f^{t}_\ell(\cdot)$ (in practice, each stage of a ResNet~\cite{he2016resnet}). As a reminder, $c$ is the channel and $w\times h$ are the spatial coordinates. All POD variants use the Euclidean distance of $\ell^2$-normalize tensors, here noted as $\left\Vert\cdot-\cdot\right\Vert$. They differ on the type of pooling applied before that distance is computed.
%
%
On one extreme, one can apply no pooling at all, resulting in the most strict loss, the most rigid constrains, and the lowest plasticity:
\begin{equation}
    \mcL_{\text{POD-pixel}}(\vh^{t-1}_\ell, \vh^t_\ell) = \sum_{c=1}^C \sum_{w=1}^{W} \sum_{h=1}^{H} \left\Vert \vh^{t-1}_{\ell,c,w,h} - \vh^t_{\ell,c,w,h} \right\Vert^2\label{eq:pod_pixel}\,.
\end{equation}
By pooling the channels, one preserves only the spatial coordinates, resulting in a more permissive loss, allowing the activations to reorganize across the channels, but penalizing global changes of those activations across the space,
\begin{equation}
    \mcL_{\text{POD-channel}}(\vh^{t-1}_\ell, \vh^t_\ell)  = \sum_{w=1}^{W} \sum_{h=1}^{H} \left\Vert \sum_{c=1}^C \vh^{t-1}_{\ell,c,w,h} - \sum_{c=1}^C \vh^{t}_{\ell,c,w,h} \right\Vert^2\label{eq:pod_channel}\,;
\end{equation}
or, contrarily, by pooling the space (equivalent, up to a factor, to a Global Average Pooling), one preserves \textit{only} the channels:
\begin{equation}
    \mcL_{\text{POD-gap}}(\vh^{t-1}_\ell, \vh^t_\ell) = \sum_{c=1}^{C} \left\Vert \sum_{w=1}^{W} \sum_{h=1}^H \vh^{t-1}_{\ell,c,w,h} - \sum_{w=1}^{W} \sum_{h=1}^H \vh^{t}_{\ell,c,w,h} \right\Vert^2\label{eq:pod_gap}\,.
\end{equation}

Note that the only difference between the variants is in the position of the summation. For example, contrast equations \autoref{eq:pod_pixel} and \ref{eq:pod_channel}: in the former the differences are computed between activation pixels, and then totaled; in the latter, first the channel axis is flattened, then the differences are computed, resulting in a more permissive loss. 

We can trade a little plasticity for rigidity, with less aggressive pooling by aggregating statistics across just one of the spatial dimensions:
\begin{equation}
    \mcL_{\text{POD-width}}(\vh^{t-1}_\ell, \vh^t_\ell)  = \sum_{c=1}^{C} \sum_{h=1}^{H} \left\Vert \sum_{w=1}^W \vh^{t-1}_{\ell,c,w,h} - \sum_{w=1}^W \vh^{t}_{\ell,c,w,h} \right\Vert^2\label{eq:pod_width}\,;
\end{equation}
%
or, likewise, for the vertical dimension, resulting in POD-height. Each of those variants measure the distribution of activation pixels across their respective axis. These two complementary intermediate statistics can be further combined together:
\begin{equation}
    \mcL_{\text{POD-spatial}}(\vh^{t-1}_\ell, \vh^t_\ell) = \mcL_{\text{POD-width}}(\vh^{t-1}_\ell, \vh^t_\ell) + \mcL_{\text{POD-height}}(\vh^{t-1}_\ell, \vh^t_\ell)\,.
\end{equation}
$\mcL_{\text{POD-spatial}}$ is minimal when the average statistics over the dataset, on both width and height axes, are similar for the previous and current model. It brings the right balance between being too rigid (\autoref{eq:pod_pixel}) and being too permissive (\autoref{eq:pod_channel} and \ref{eq:pod_gap}).

\label{sec:pod_flat}
\parag{Constraining the final embedding} After the convolutional layers, the network, by design, flattens the spatial coordinates, and the formalism above needs adjustment, as a summation over $w$ and $h$ is no longer possible. Instead, we set a flat constraint on the final embedding $\vh^{t} = f^{t}(\vx)$:
\begin{equation}
    \mcL_{\text{POD-flat}}(\vh^{t-1}, \vh^t) = \left\Vert \vh^{t-1} - \vh^t \right\Vert^2\label{eq:POD-flat}\,.
\end{equation}

\parag{Combining the losses, analysis} The final POD loss combines the two  components:
\begin{multline}
    \mcL_\text{POD-final}(\vx) =  \frac{\lambda_{c}}{L-1}\sum_{\ell=1}^{L-1}  \mcL_{\text{POD-spatial}}\left(f^{t-1}_\ell(\vx), f^t_\ell(\vx)\right) + \\[-0.8em]
    \lambda_{f} \mcL_\text{POD-flat}\left(f^{t-1}(\vx), f^t(\vx)\right)\,.
\end{multline}
The hyperparameters $\lambda_{c}$ and $\lambda_{f}$ are necessary to balance the two terms, due to the  different nature of the intermediate outputs (spatial and flat).

As mentioned, the strategy above generalizes disparate propositions existing both in the literature of incremental learning, and elsewhere. When $\lambda_{c}=0$, it reduces to the cosine constraint of \textit{Less-Forget}, proposed by Hou et al. for incremental learning, which constrains only the final embedding~\cite{hou2019ucir}. When $\lambda_{f}=0$ and POD-spatial is replaced by POD-pixel, it suggests the Perceptual Features loss, proposed for style transfer~\cite{johnson2016perceptual_losses}. When $\lambda_{f}=0$ and POD-spatial is replaced by POD-channel, the strategy hints at the loss proposed by Komodakis et al.~\cite{komodakis2017attention_residual_distillation} to allow distillation across different networks, a situation in which the channel pooling responds to the very practical need to allow the comparison of architectures with different number of channels.

As we will see in our evaluations of pooling strategies (\autoref{sec:ablation_pooling}), what proved optimal was a completely novel idea, POD-spatial, combining two poolings, each of which flattens one of the spatial coordinates. That relatively rigid strategy (channels and one of the spatial coordinates are considered in each half of the loss) makes intuitive sense in our context, which is \textit{small-task} incremental learning, and thus where we expect a slow drift of the model across a single task.


\subsection{Local Similarity Classifier}
\label{sec:clf}

Hou et al.~\cite{hou2019ucir} observed that the class imbalance of incremental learning have concrete manifestations on the parameters of the final layer on classifiers, namely the weights for the over-represented (new) classes becoming much larger than those for the underrepresented (old) classes. To overcome this issue, their method (called here UCIR) $\ell^2$-normalizes both the weights and the activations, which corresponds to taking the cosine similarity instead of the dot product. For each class $c$, their last layer becomes
\begin{equation}
\vyh_{c}=\frac{\exp\left(\eta\langle\theta_{c},\vh\rangle\right)}{\sum_{i} \exp \left(\eta\langle\theta_{i}, \vh\rangle\right)}\,,
\end{equation}
where $\theta_c$ are the last-layer weights for class $c$, $\eta$ is a learned scaling parameter, and $\langle\cdot,\cdot\rangle$ is the cosine similarity.

However, this strategy optimizes a \textit{global similarity}: its training objective increases the similarity between the extracted features and their associated weights. For each class, the normalized weight vector acts as a \textit{single} proxy~\cite{attias2017proxynca}, towards which the learning procedure pushes all samples in the class.

We observed that such global strategy is hard to optimize in an incremental setting. To avoid forgetting, the distillation losses (\autoref{sec:distillation}) tries to keep the final embedding $\vh$ consistent through time so that the class proxies stay relevant for the classifier. Unfortunately catastrophic forgetting, while alleviated by current methods, is not solved and thus the distribution of $\vh$ may change. The cosine classifier is very sensitive to those changes as it models a unique majority mode through its class proxies.

\parag{Local Similarity Classifier} The problem above lead us to amend the classification layer during training, in order to consider multiple proxies/modes per class. A shift in the distribution of $\vh$ will have less impact on the classifier as more modes are covered.


Our redesigned classification layer, which we call Local Similarity Classifier (LSC), allows for $K$ multiple proxies/modes during training. Like before, the proxies are a way to interpret the weight vector in the cosine similarity, thus we allow for $K$ vectors $\theta_{c,k}$ for each class $c$.
The similarity $s_{c,k}$ to each proxy/mode is first computed. An averaged class similarity $\vyh_c$ is the output of the classification layer:
\begin{equation}
s_{c,k} =\frac{\exp\,\langle\theta_{c,k},\vh\rangle}{\sum_{i} \exp\,\langle\theta_{c,i},\vh\rangle}\,, \qquad
\vyh_c = \sum_{k}s_{c,k}\,\langle\theta_{c,k},\vh\rangle\,.
\end{equation}
The multi-proxies classifier optimizes the similarity of each sample to its ground truth class representation and minimizes all others. A simple cross-entropy loss would work, but we found empirically that the NCA loss~\cite{goldberger2005nca_loss,attias2017proxynca} converged faster. We added to the original loss a hinge $[\,\cdot\,]_+$ to keep it bounded, and a small margin $\delta$ to enforce stronger class separation, resulting in the final formulation:
\begin{equation}
    \mcL_\text{LSC} = \left[- \log\frac{\exp\left(\eta (\vyh_y - \delta)\right)}{\sum_{i \neq y} \exp \eta \vyh_{i}} \right]_+ \,.
\end{equation}

\parag{Weight initialization for new classes} The incremental learning setting imposes detecting new classes at each new task $t$. New weights $\{\theta_{c,k} \mid \forall c \in C^t_N, \forall k \in {1...K}\}$ must be added to predict them. We could initialize them randomly, but the class-agnostic features of the ConvNet $f$, extracted by the model trained so far offer a better prior. Thus, we employ a generalization of Imprinted Weights~\cite{qi2018imprintedweights} procedure to multiple modes: for each new class $c$, we extract the features of its training samples, use a k-means algorithm to split them into $K$ clusters, and use the centroids of those clusters as initial values for $\theta_{c,k}$. This procedure ensures mode diversity at the beginning of a new task and resulted in a one percentage point improvement on CIFAR100 \cite{krizhevskycifar100}.


\subsection{Complete model formulation}

Our model has the classical structure of a convolutional network $f(\cdot)$ acting as a features extractor, and a classifier $g(\cdot)$ producing a score per class. We introduced two innovations to this model: (1) our main contribution is a novel distillation loss (POD) applied all over the ConvNet, from the spatial features $\vh_\ell$ to the final flat embedding $\vh$; (2) as further refinement we propose that the classifier learns a multi-modal representation that explicitly keeps multiple proxy vectors per class, increasing the model expressiveness and thus making it less sensible to shift in the distribution of $\vh$. The final loss for current model $g^t \circ f^t$, i.e., the model trained for task $t$, is simply their addition $\mathcal{L}_{\{f^t; g^t\}} = \mathcal{L}_\textrm{LSC} + \mathcal{L}_\textrm{POD-final}$.
%

\section{Experiments}
\label{sec:expes}



We compare our technique (PODNet) with three state-of-the-art models. Those models are particularly comparable to ours since they all employ a sample memory with a fixed capacity. Both iCaRL~\cite{rebuffi2017icarl} and UCIR~\cite{hou2019ucir} use the same inference method --\,\textit{Nearest-Mean-Examplars} (NME), although UCIR also proposes a second inference method based on the classifier probabilities (called here UCIR-CNN). We evaluate PODNet with both inference methods for a small scale dataset, and the later for larger scale datasets. BiC~\cite{wu2019bias_correction}, while not focused on representation learning, is specially designed to be effective on large scale datasets, and thus provided an interesting baseline.

\parag{Datasets} We employ three images datasets --\,extensively used in the literature of incremental learning\,-- for our experiments: CIFAR100~\cite{krizhevskycifar100}, ImageNet100~\cite{deng2009imagenet,hou2019ucir,wu2019bias_correction}, and ImageNet1000~\cite{deng2009imagenet}.
ImageNet100 is a subset of ImageNet1000 with only 100 classes, randomly sampled from the original 1000.


\parag{Protocol} We validate our model and the compared baselines using the challenging protocol introduced by Hou et al.~\cite{hou2019ucir}: we start by training the models on half the classes (i.e., 50 for CIFAR100 and ImageNet100, and 500 for ImageNet1000). Then the classes are added incrementally in steps. We divide the remaining classes equally among the steps, e.g., for CIFAR100 we could have 5 steps of 10 classes or 50 steps of 1 class. Note that a training of 50 steps is actually made of 51 different tasks: the initial training followed by the incremental steps. Models are evaluated after each step on \textit{all the classes seen until then}. To facilitate comparison, the accuracies at the end of each step are averaged into a unique score called \textit{average incremental accuracy}~\cite{rebuffi2017icarl}. If not specified otherwise, the average incremental accuracy is the score reported in all our results.


Following Hou et al.~\cite{hou2019ucir}, for all datasets, and all compared models, we limit the memory $M_\textrm{per}$ to 20 images per old class. For results with different memory settings, refer to \autoref{sec:robustness}.

\parag{Implementation details} For fair comparison, all compared models employ the same ConvNet backbone: ResNet-32 for CIFAR100, and ResNet-18 for ImageNet. We remove the ReLU activation at the last block of each ResNet end-of-stage to provide a signed input to POD (\autoref{sec:distillation}). We implemented our method (called here PODNet) in PyTorch~\cite{paszke2017pytorch}. 
We compare both ours and UCIR's implementation~\cite{hou2019ucir} of iCaRL. Results of UCIR come from the implementation of Hou et al.~\cite{hou2019ucir}. We provide their reported results and also run their code ourselves. We used our implementation of BiC in order to compare with the same backbone.
We sample our memory images using \textit{herding selection}~\cite{rebuffi2017icarl} and perform the inference with two different methods: the \textit{Nearest-Mean-Examplars} (NME) proposed for iCarl, and also adopted on one of the variants of UCIR~\cite{hou2019ucir}, and the ``CNN'' method introduced for UCIR (see \autoref{sec:related_work}).
Please see the supplementary materials for the full implementation details.

\begin{table*}[t]
\caption{Average incremental accuracy for PODNet \textit{vs.} state of the art. We run experiments three times (random class orders) on CIFAR100 and report averages\mypm{}standard deviations. Models with an asterisk * are reported directly from Hou et al~\cite{hou2019ucir}}
\label{tab:quantitative_cifar}
\centering
\begin{tabular}{@{}l|cccc@{}}
 \toprule
 & \multicolumn{4}{c}{CIFAR100}\\
  & 50 steps & 25 steps & 10 steps & 5 steps\\
 \multicolumn{1}{r|}{New classes per step} & 1 & 2 & 5 & 10\\
 \midrule
 \textit{iCaRL*} \cite{rebuffi2017icarl} & --- & --- & 52.57$\mspace{51mu}$ & 57.17$\mspace{51mu}$\\
 iCaRL & 44.20\mypm{}0.98    & 50.60\mypm{}1.06  & 53.78\mypm{}1.16  & 58.08\mypm{}0.59\\
 BiC \cite{wu2019bias_correction} & 47.09\mypm{}1.48 & 48.96\mypm{}1.03 & 53.21\mypm{}1.01  & 56.86\mypm{}0.46\\
 \textit{UCIR\,{\scriptsize (NME)}*} \cite{hou2019ucir} & --- & --- & 60.12$\mspace{51mu}$ & 63.12$\mspace{51mu}$\\
 UCIR\,{\scriptsize (NME)} \cite{hou2019ucir}  & 48.57\mypm{}0.37 & 56.82\mypm{}0.19 & 60.83\mypm{}0.70 & 63.63\mypm{}0.87\\
 \textit{UCIR\,{\scriptsize (CNN)}*} \cite{hou2019ucir}  & --- & --- & 60.18$\mspace{51mu}$ & 63.42$\mspace{51mu}$\\
 UCIR\,{\scriptsize (CNN)} \cite{hou2019ucir} & 49.30\mypm{}0.32 & 57.57\mypm{}0.23 & 61.22\mypm{}0.69 & 64.01\mypm{}0.91\\
 PODNet\,{\scriptsize (NME)} & \textbf{61.40\mypm{}0.68} & \textbf{62.71\mypm{}1.26} & \textbf{64.03\mypm{}1.30} & \textbf{64.48\mypm{}1.32}\\
 PODNet\,{\scriptsize (CNN)} & \textbf{57.98\mypm{}0.46} & \textbf{60.72\mypm{}1.36} & \textbf{63.19\mypm{}1.16} & \textbf{64.83\mypm{}0.98}\\
 \bottomrule
\end{tabular}
\end{table*}

\begin{table*}[t]
\caption{Average incremental accuracy, PODNet \textit{vs.} state of the art. Models with an asterisk * are reported directly from Hou et al.~\cite{hou2019ucir}}
\label{tab:quantitative_imagenet}
\centering
\begin{tabular}{@{}l|cccc|cc@{}}
 \toprule
 & \multicolumn{4}{c|}{ImageNet100} & \multicolumn{2}{c}{Imagenet1000} \\
  & 50 steps & 25 steps & 10 steps & 5 steps & 10 steps & 5 steps\\
 \multicolumn{1}{r|}{New classes per step} & 1 & 2 & 5 & 10 & 50 & 100\\
 \midrule
 iCaRL* \cite{rebuffi2017icarl}        & ---   & ---   & 59.53  & 65.04 & 46.72 & 51.36\\
 iCaRL \cite{rebuffi2017icarl}         & 54.97 & 54.56 & 60.90  & 65.56 & ---   & --- \\
 BiC \cite{wu2019bias_correction} & 46.49 & 59.65 & 65.14  & 68.97 & 44.31   & 45.72\\
 UCIR\,{\scriptsize (NME)}* \cite{hou2019ucir}    & ---   & ---   & 66.16  & 68.43 & 59.92 & 61.56\\
 UCIR\,{\scriptsize (NME)} \cite{hou2019ucir}     & 55.44 & 60.81 & 65.83  & 69.07 & ---   & --- \\
 UCIR\,{\scriptsize (CNN)}* \cite{hou2019ucir}    & ---   & ---   & 68.09  & 70.47 & 61.28 & 64.34\\
 UCIR\,{\scriptsize (CNN)} \cite{hou2019ucir}     & 57.25 & 62.94 & 67.82  & 71.04 & ---   & --- \\
 PODNet\,{\scriptsize (CNN)}                & \textbf{62.48} & \textbf{68.31} & \textbf{74.33} & \textbf{75.54} & \textbf{64.13} & \textbf{66.95}\\

  & \scriptsize{\textbf{$\pm$ 0.59}} & \scriptsize{\textbf{$\pm$ 2.45}} & \scriptsize{\textbf{$\pm$ 0.93}} & \scriptsize{\textbf{$\pm$ 0.26}} & &\\

 \bottomrule
\end{tabular}
\end{table*}

\subsection{Quantitative results}
\label{sec:quantitative_results}

The comparisons with all the state of the art are tabulated in \autoref{tab:quantitative_cifar} for CIFAR100 and \autoref{tab:quantitative_imagenet} for ImageNet100 and ImageNet1000. All tables shows the average incremental accuracy for each considered models with various number of steps on the incremental learning run. The ``New classes per step'' row shows the amount of new classes introduced per task. 

\begin{figure*}[tb]
  \centering
  \begin{subfigure}[b]{0.48\linewidth}
    \includegraphics[width=\linewidth]{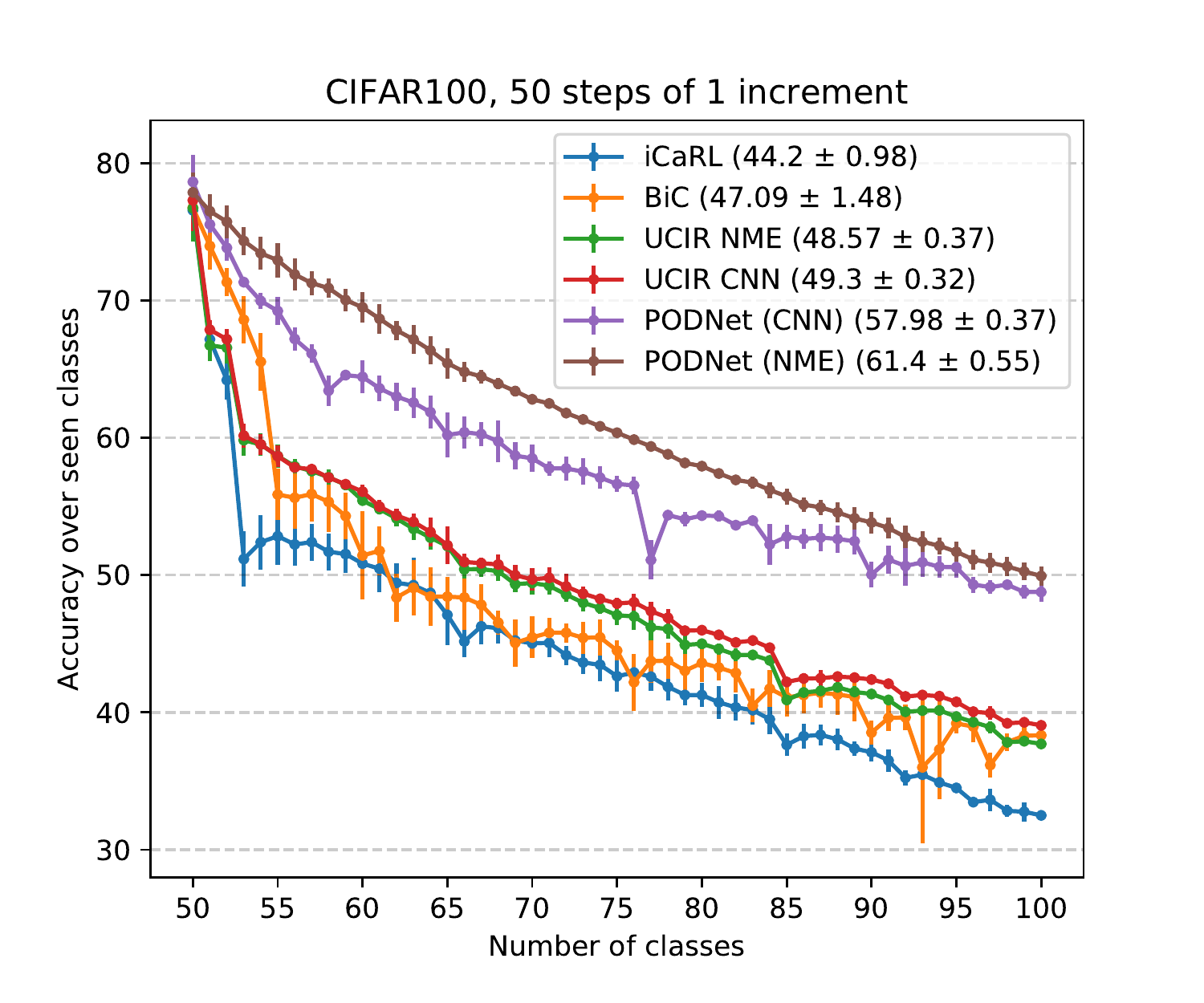}
    \caption{50 steps, 1 class / step}
    \label{fig:cifar_inc1}
  \end{subfigure}
  \hfill
  \begin{subfigure}[b]{0.48\linewidth}
    \includegraphics[width=\linewidth]{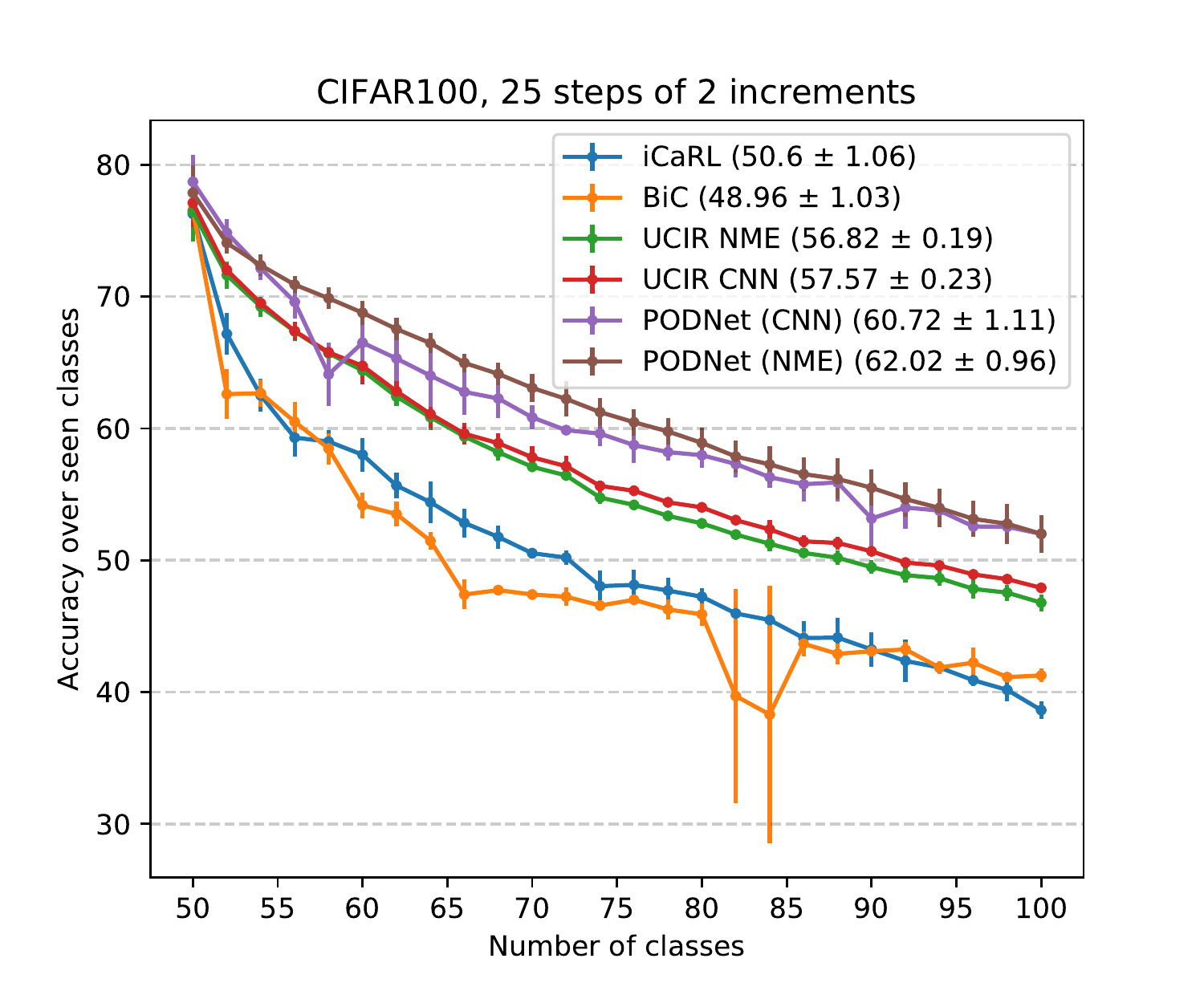}
    \caption{25 steps, 2 classes / step}
    \label{fig:cifar_inc2}
  \end{subfigure}
  \caption{\textbf{Incremental Accuracy on CIFAR100} over three orders for two different step sizes. The legend reports the average incremental accuracy.}
  \label{fig:plots}
\end{figure*}

\parag{CIFAR100} We run our comparisons on 5, 10, 25, and 50 steps with respectively 10, 5, 2, and 1 classes per step. We created three random class orders to ran each experiment thrice, reporting averages and standard deviations. For CIFAR100 only, we evaluated our model with two different kind of inference: NME and CNN. With both methods, our model surpasses all previous state of the art models on all steps. Moreover, our model relative improvement grows as the number the steps increases, surpassing existing models by 0.82, 2.81, 5.14, and 12.1 percent points (\pp) for respectively 5, 10, 25, and 50 steps. Larger numbers of steps imply  stronger forgetting; those results confirm that PODNet manages to reduce drastically the said forgetting. While PODNet with NME has the largest gain, PODNet with CNN also outperforms the previous state of the art by up to 8.68\pp. See \autoref{fig:plots} for a plot of the incremental accuracies on this dataset. In the extreme setting of 50 increments of 1 class (\autoref{fig:cifar_inc1}), our model showcases large differences, with slow degradation (``\textit{gradual forgetting}'' \cite{french1999catastrophicforgetting}) due to forgetting throughout the run, while the other models show a quick performance collapse (``\textit{catastrophic forgetting}'') at the start of the run.

\parag{ImageNet100} We run our comparisons on 5, 10, 25, and 50 steps with respectively 10, 5, 2, and 1 classes per step. For both ImageNet100, and ImageNet1000 we report only PODNet with CNN, as the kNN-based NME classifier did not generalize as well to larger-scale datasets. With the more complex images of ImageNet100, our model also outperforms the state of the art on all tested runs, by up to 6.51\pp.

\parag{ImageNet1000} This dataset is the most challenging, with much greater image complexity than CIFAR100, and ten times the number of classes as ImageNet100. We evaluate the models in 5 and 10 steps, and results confirm the consistent improvement of PODNet against existing arts by up to 2.85\pp. 

\captionsetup[table]{skip=0pt}
\begin{table}[tb]
\caption{Ablation studies performed on CIFAR100 with 50 steps. We report the average incremental accuracy}
\setlength{\tabcolsep}{3.2pt}
\begin{subtable}[t]{.57\textwidth}
\centering
\caption{Comparison of the performance of the model when disabling parts of the complete PODNet loss\\~}
\label{tab:ablation_inc}
\begin{tabular}{@{}lccccc@{}}
 \toprule
 Classifier & POD-flat & POD-spatial & NME & CNN\\
 \midrule
 Cosine     &     &     & 40.76 & 37.93\\
 Cosine     & \OK &     & 48.03 & 46.73 \\
 Cosine     &     & \OK & 54.32 & 57.27 \\
 Cosine     & \OK & \OK & 56.69 & 55.72 \\
 LSC-CE     & \OK & \OK & 59.86 & 57.45 \\
 LSC        &     &     & 41.56 & 40.76 \\
 LSC        & \OK &     & 53.29 & 52.98 \\
 LSC        &     & \OK & \textbf{61.42} & 57.64 \\
 LSC        & \OK & \OK & 61.40 & \textbf{57.98} \\
 \bottomrule
\end{tabular}
\end{subtable}
\hfill
\begin{subtable}[t]{.42\textwidth}
\centering
\caption{Comparison of distillation losses based on intermediary features. All losses evaluated with POD-flat}
\label{tab:ablation_perceptual}
\begin{tabular}{@{}lcc@{}}
 \toprule
 Loss      & NME & CNN\\
 \midrule
 \textit{None}                  & 53.29  & 52.98\\
 POD-pixels                     & 49.74  & 52.34 \\
 POD-channels                   & 57.21  & 54.64\\
 POD-gap                        & 58.80  & 55.95\\
 POD-width                      & 60.92  & 57.51\\
 POD-height                     & 60.64  & 57.50\\
 POD-spatial                    & \textbf{61.40} & \textbf{57.98}\\
 \cmidrule{1-3}
 GradCam~\cite{dhar2019learning_without_memorizing_gradcam}              & 54.13 & 52.48 \\
 Perceptual Style~\cite{johnson2016perceptual_losses}       & 51.01 & 52.25  \\
 \bottomrule
\end{tabular}
\end{subtable}

\end{table}
\captionsetup[table]{skip=10pt}

\subsection{Further analysis \& ablation studies}
\label{sec:ablation}

\parag{Ablation Studies}
Our model has two components: the distillation loss POD and the LSC classifier. An ablation study showcasing the contribution of each component is displayed in \autoref{tab:ablation_inc}: each additional component improves the model performance. We evaluate every ablation on CIFAR100 with 50 steps of 1 new class each. The reported metric is the average incremental accuracy. The table shows that our novel method of constraining the whole ConvNet is beneficial. Furthermore applying only POD-spatial still beats the previous state of the art by a significant margin. Using both POD-spatial and POD-flat then further increases results with a large gain. We also compare the results with the Cosine classifier~\cite{luo2018cosine_classifier,hou2019ucir} against the Local Similarity Classifier (LSC) with NCA loss. Finally, we add LSC-CE: our classifier with multi-mode but with a simple cross-entropy loss instead of our modified NCA loss. This version brings to mind SoftTriple~\cite{qian2019softtriple} and Infinited Mixture Prototypes~\cite{allen2019infinitemixtureproto}, used in the different context of few-shot learning.
The latter only considers the closest mode of each class in its class assignment, while LSC considers all modes of a class, thus, taking into account the intra-class variance. That allows LSC to decrease class similarity when intra-class variance is high (which could signal a lack of confidence in the class).

\label{sec:ablation_pooling}
\parag{Spatial-based distillation} We apply our distillation loss POD differently for the flat final embedding $\vh$ (POD-flat) and the ConvNet's intermediate features maps $\vh_\ell$ (POD-spatial). We designed and evaluated several alternative for the latter whose results are shown in \autoref{tab:ablation_perceptual}. Refer to \autoref{sec:distillation} and \autoref{fig:pooling} for their definition. All losses are evaluated with POD-flat. "\textit{None}" is using only POD-flat.
Overall, we see that not using pooling results in bad performance (POD-pixels). Our final loss, POD-spatial, surpasses all others by taking advantages of the statistics aggregated from both spatial axis. For the sake of completness we also included losses not designed by us: GradCam distillation~\cite{dhar2019learning_without_memorizing_gradcam} and Perceptual Style~\cite{johnson2016perceptual_losses}. The former uses a gradient-based attention while the later --\,used for style transfer\,-- computes a gram matrix for each channel.

\parag{Forgetting and plasticity balance} Forgetting can be drastically reduced by imposing a high factor on the distillation losses. Unfortunately, it will also degrade the capacity (its \textit{plasticity}) to learn new classes. When POD-spatial is added on top of POD-flat, we manage to increase the oldest classes performance (+7 percentage points) while the newest classes performance were barely reduced (-0.2\pp). Because our loss POD-spatial constraints only statistics, it is less stringent than a loss based on exact pixels values as POD-pixel. The latter hurts the newest classes (-2\pp) for a smaller improvement of old classes (+5\pp). Furthermore our experiments confirmed that LSC reduced the sensibility of the model to distribution shift, as the performance it brings was localized on the old classes.

\label{sec:robustness}
\parag{Robustness of our model} While previous results showed that PODNet improved significantly over the state-of-the-arts, we wish here to demonstrate here the robustness of our model to various factors. In \autoref{tab:ablation_memorysize}, we compared how PODNet behaves against the baseline when the memory size per class $M_{\text{per}}$ changes: PODNet improvements increase as the memory size decrease, up to a gain of 26.20\pp\ with NME (resp. 13.42\pp\ for CNN) with $M_{\text{per}} = 5$. Notice that by default, the memory size is 20 in \autoref{sec:quantitative_results}.
We also compared our model against baselines with a more flexible memory $M_{\text{total}} = 2000$ \cite{rebuffi2017icarl,wu2019bias_correction}, and with various initial task size (by default it is 50 on CIFAR100). In the former case (\autoref{tab:sub_free_memory}), models benefit from a larger memory per class in the early tasks. In the later case (\autoref{tab:sub_initialincrement}), models initialization is worse because of a smaller initial task size. In these settings very different from \autoref{sec:quantitative_results}, PODNet still outperformed significantly the compared models, proving the robustness of our model.

\begin{table}[t]
\centering
\caption{Effect of the memory size per class $M_{per}$ on the models performance. Results from CIFAR100 with 50 steps, we report the average incremental accuracy}
\label{tab:ablation_memorysize}
\begin{tabular}{@{}lccccccc@{}}
 \toprule
 $M_{per}$ & 5     & 10    & \textbf{20}    & 50    & 100   & 200\\
 \midrule
iCaRL \cite{rebuffi2017icarl}      & 16.44 & 28.57 & 44.20 & 48.29 & 54.10 & 57.82\\
BiC \cite{wu2019bias_correction}       & 20.84  & 21.97  & 47.09  & 55.01  & 62.23  & \textbf{67.47}\\
UCIR\,{\scriptsize (NME)} \cite{hou2019ucir} & 21.81 & 41.92 & 48.57 & 56.09 & 60.31 & 64.24\\
UCIR\,{\scriptsize (CNN)} \cite{hou2019ucir} & 22.17 & 42.70 & 49.30 & 57.02 & 61.37 & 65.99\\
PODNet\,{\scriptsize (NME)}& \textbf{48.37} & \textbf{57.20} & \textbf{61.40} & \textbf{62.27} & \textbf{63.14} & 63.63\\
PODNet\,{\scriptsize (CNN)} & \textbf{35.59} & \textbf{48.54} & \textbf{57.98} & \textbf{63.69} & \textbf{66.48} & \textbf{67.62} \\
\bottomrule
\end{tabular}
\end{table}

\captionsetup[table]{skip=0pt}
\begin{table}[t]
\setlength{\tabcolsep}{2.7pt}
\caption{Effect of the initial task size and the $M_\mathrm{total}$ on the models performance. We report the average incremental accuracy}
\begin{subtable}[b]{.36\textwidth}
\centering
\caption{Evaluation of an easier memory constraint ($M_\mathrm{total} = 2000$)}
\label{tab:sub_free_memory}
\begin{tabular}{@{}lcc@{}}
 \toprule
           & \multicolumn{2}{c}{Nb. steps} \\
           \cmidrule{2-3}
 Loss      & 50 & 10 \\
 \midrule
iCaRL \cite{rebuffi2017icarl}  & 42.34 & 56.52\\
BiC \cite{wu2019bias_correction} & 48.44 & 55.03\\
UCIR\,{\scriptsize (NME)}\,\cite{hou2019ucir} & 54.08 & 62.89\\
UCIR\,{\scriptsize (CNN)}\,\cite{hou2019ucir} & 55.20 & 63.62\\
PODNet\,{\scriptsize (NME)} & \textbf{62.47} & \textbf{64.60}\\
PODNet\,{\scriptsize (CNN)} & \textbf{61.87} & \textbf{64.68}\\
 \bottomrule
\end{tabular}
\end{subtable}
\hfill
\begin{subtable}[b]{.62\textwidth}
\centering
\caption{Varying initial task size, with $M_\mathrm{per} = 20$, and followed by 50 to 90 tasks made of a single class}
\label{tab:sub_initialincrement}
\begin{tabular}{@{}lccccc@{}}
 \toprule
      & \multicolumn{5}{c}{Initial task size} \\
      \cmidrule{2-6}
 Loss & 10 & 20 & 30 & 40 & \textbf{50}\\
 \midrule
iCaRL \cite{rebuffi2017icarl}     & 40.97 & 41.28 & 43.38 & 44.35 & 44.20\\
BiC \cite{wu2019bias_correction}       & 41.58 & 40.95  & 42.27 & 45.18 & 47.09\\
UCIR\,{\scriptsize (NME)} \cite{hou2019ucir} & 42.33 & 40.81 & 46.80 & 46.71 & 48.57\\
UCIR\,{\scriptsize (CNN)} \cite{hou2019ucir} & 43.25 & 41.69 & 47.85 & 47.51 & 49.30\\
PODNet\,{\scriptsize (NME)}& \textbf{45.09} & \textbf{49.03} & \textbf{55.30} & \textbf{57.89} & \textbf{61.40}\\
PODNet\,{\scriptsize (CNN)}& \textbf{44.95} & \textbf{47.68} & \textbf{52.88} & \textbf{55.42} & \textbf{57.98}\\
\bottomrule
\end{tabular}
\end{subtable}
\end{table}
\captionsetup[table]{skip=10pt}

\section{Conclusion}

We introduced in this paper a novel distillation loss (POD) constraining the whole convolutional network. This loss strikes a balance between reducing forgetting of old classes and learning new classes, essential for long incremental runs, by carefully chosen pooling. As a further refinement, we proposed a multi-mode similarity classifier, more robust to shift in the distribution inherent to incremental learning.
Those innovations allow PODNet to outperform the previous state of the art in a challenging experimental context, with severe sample-per-class memory limitation, and long runs of many small-sized tasks, by a large margin. Extensive experiments over three datasets show the robustness of our model on different settings.

\parag{Acknowledgement} E. Valle is funded by FAPESP grant 2019/05018-1 and CNPq grants 424958/2016-3 and 311905/2017-0. This work was performed using HPC resources from GENCI–IDRIS (Grant 2020-AD011011706). We also wish to thanks Estelle Thou for the helpful discussion.

%
%

\section{Supplementary Material}

\subsection{Spatial-based distillation without POD-flat}

In Table 3.b of the main paper, we compared  distillation loss alternatives to our final POD-spatial. In this table, the spatial-based losses were evaluated with POD-flat. In \autoref{tab:ablation_perceptual_noflat}, we evaluate those same losses without POD-flat. "\textit{None}" is using only our LSC classifier without any distillation losses. Notice that POD-spatial ---and its sub-components POD-width and POD-height-- are the only losses barely affected by POD-flat's absence. Note that all alternative losses were tuned on the validation set to get the best performance, including those from external papers. Still, our proposed loss, POD-spatial, outperforms all, both with and without POD-flat.

\begin{table*}[!htbp]
\centering
\caption{Comparison of distillation losses based on intermediary features. All losses evaluated without POD-flat.}
\begin{tabular}{@{}lcc@{}}
 \toprule
 Loss      & NME & CNN\\
 \midrule
 \textit{None}                  & 41.56  & 40.76\\
 POD-pixels                     & 42.21  & 40.81 \\
 POD-channels                   & 55.91  & 50.34\\
 POD-gap                        & 57.25  & 53.87\\
 POD-width                      & 61.25  & 57.51\\
 POD-height                     & 61.24  & 57.50\\
 POD-spatial                    & \textbf{61.42} & \textbf{57.64}\\
 \hdashline
 GradCam~\cite{dhar2019learning_without_memorizing_gradcam}              & 41.89 & 42.07 \\
 Perceptual Style~\cite{johnson2016perceptual_losses}       & 41.74 & 40.80 \\
 \bottomrule
\end{tabular}
\label{tab:ablation_perceptual_noflat}
\end{table*}

\subsection{Implementation details}

For all datasets, images are augmented with random crops and flips. For CIFAR100, we additionally change image intensity by a random value in the range [-63, 63].
We train our model for 160 epochs for CIFAR100, and 90 epochs for both ImageNet100 and ImageNet100, with a SGD optimizer with momentum of 0.9. For all datasets, we start with a learning rate of 0.1, a batch size of 128, and cosine annealing scheduling.
The weight decay is $5\cdot 10^{-4}$ for CIFAR100, and $1\cdot 10^{-4}$ for ImageNet100 and ImageNet1000. For CIFAR100 we set model hyperparameters $\lambda_c = 3$ and $\lambda_f=1$, while for ImageNet100 and 1000 we set $\lambda_c = 8$ and $\lambda_f =10$. Our model uses POD-spatial and POD-flat except when explicitly stated otherwise. Following Hou et al.~\cite{hou2019ucir}, we multiply both losses by the adaptive scaling factor: $\lambda=\sqrt{\nicefrac{N}{T}}$ with $N$ being the number of seen classes and $T$ the number of classes in the current task.

For POD-spatial, before sum-pooling we take the features to the power of 2 element-wise. The vector resulting from the pooling is then L2 normalized. 

\subsection{Number of proxies per class}

While our model's expressiveness increases with more proxies in $\mcL_\text{LSC}$, it remains fairly  stable for values between 5 and 15, thus, for simplicity, we kept it fixed to 10 in all experiments.

In initial experiments, we had the following pairs for the number of clusters (k) and average incremental accuracy (acc): k=1, acc=56.80\%; k=2, 57.14\%; k=4, acc=57.40\%; k=6, acc=57.46\%; k=8, acc=57.95\%, and k=10, acc=57.98\% --- i.e., a 1.18 p.p. improvement moving from k=1 to k=10. On ImageNet100, with 10 steps/tasks (increments of give classes per task), moving from k=1 to k=10 improved 1.51 p.p. on acc.

\subsection{Reproducibility}

\textbf{Code Dependencies} The Python version is  3.7.6. We used the PyTorch \cite{paszke2017pytorch} (version 1.2.0) deep learning framework and the libraries Torchvision (version 0.4.0), NumPy \cite{oliphant2006numpy} (version 1.17.2), Pillow (version 6.2.1), and Matplotlib \cite{hunter2007matplotlib} (version 3.1.0). The CUDA version is 10.2. Initial experiments were done with the data loaders library Continuum \cite{douillardlesort2020continuum}. PODNet's full code is released at:\\ \href{https://github.com/arthurdouillard/incremental\_learning.pytorch}{\texttt{github.com/arthurdouillard/incremental\_learning.pytorch}}. \\We provide all configuration files necessary to reproduce results, including seeds and class ordering.


\bibliographystyle{splncs04}
\bibliography{egbib}

\end{document}